\documentclass[letterpaper, 10pt, conference]{ieeeconf} 
\makeatletter
\let\NAT@parse\undefined
\makeatother

\usepackage{dblfloatfix}
\usepackage[numbers, sectionbib, sort]{natbib}
\usepackage{bm}
\usepackage{gensymb}
\usepackage{xcolor}
\usepackage{graphicx}
\usepackage{amsmath}
\usepackage{amssymb}
\usepackage{subcaption}
\usepackage{amsfonts}
\usepackage{multicol}
\usepackage{siunitx}
\usepackage{booktabs}
\usepackage{makecell}
\usepackage{multirow}
\usepackage{upgreek}
\usepackage[font=small]{caption}
\usepackage[export]{adjustbox}
\usepackage{tikz}
\usepackage{tabularx}
\usepackage{sidecap} \sidecaptionvpos{figure}{c}
\usepackage[hidelinks]{hyperref}
\usepackage[nameinlink, capitalize]{cleveref}
\usepackage[printonlyused,withpage,nolist,nohyperlinks]{acronym}
\usepackage{float} 
\usepackage{afterpage}  
\usepackage{placeins} 
\usepackage[T1]{fontenc}
\usepackage{balance}

\usepackage{cuted}
\usepackage[ruled,linesnumbered]{algorithm2e}
\SetNlSty{}{}{:}     
\SetKwInput{KwReq}{Require}

\Crefname{section}{Sec.}{Sec.}
\Crefname{equation}{Eq.}{Eq.}

\begin{document}
\title{\LARGE \bf
Privacy-Preserving Semantic Segmentation from High-Resolution Depth and Ultra-Low-Resolution RGB
\author{
Xuying Huang$^\star$  \and Swithinraj Moses Daniel$^\star$ \and Sicong Pan \and Sebastian Houben \and Maren Bennewitz
}
}

\twocolumn[{%
\renewcommand\twocolumn[1][]{#1}%

\maketitle

\vspace{-0.7cm}

\begin{figure}[H]
\hsize=\textwidth
\centering
\includegraphics[width=0.9\textwidth]{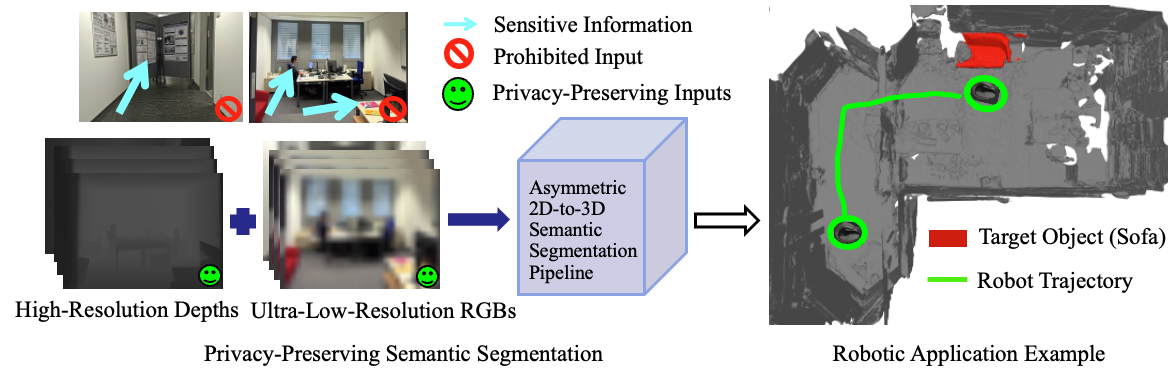}
\caption{
\textbf{Left:} During indoor mapping, high-resolution RGB observations may expose privacy-sensitive content, such as documents, screens, or human faces, and are therefore treated as prohibited inputs in privacy-preserving semantic segmentation. 
We instead use only \(640\times480\) high-resolution depth and source-level \(16\times12\) ultra-low-resolution RGB; the ultra-low-resolution images are enlarged here solely for visualization.
These highly asymmetric inputs are processed by our unified 2D-to-3D semantic segmentation pipeline to construct a semantic 3D scene representation.
\textbf{Right:} The resulting 3D semantics provide actionable information for downstream robotic tasks.
We show semantic object-goal navigation as a case study, with a sofa shown here as an example target that the robot navigates toward.
} 
\label{fig:teaser}
\end{figure}
}]

{%
\renewcommand\thefootnote{}%
\footnotetext{%
$^\star$These authors contributed equally to this work.\\
X. Huang, S. Pan, and M. Bennewitz are with the Humanoid Robots Lab, University of Bonn, the Lamarr Institute for Machine
Learning and Artificial Intelligence and the Center for Robotics, Bonn, Germany. 
S. Moses Daniel, and S. Houben are with the Bonn-Rhein-Sieg University of
Applied Sciences, Germany. S. Moses Daniel is additionally with the Humanoid
Robots Lab at the University of Bonn. S. Houben is additionally with the
Fraunhofer Institute for Intelligent Analysis and Information Systems.
This work has been partially funded by the
German Federal Ministry of Research, Technology and Space~(BMFTR) under grant No. 16KIS1949 and the Robotics Institute Germany (RIG). Corresponding: \texttt{huang@cs.uni-bonn.de}.}
}%
\setcounter{footnote}{0}

\begin{abstract}
As mobile robots become increasingly integrated into everyday environments, privacy risks arising from onboard cameras have become a growing concern.
Ultra-low-resolution (ULR) RGB can mitigate visual privacy exposure at the source, but ULR appearance alone substantially limits semantic and spatial understanding.
We therefore introduce a privacy-preserving asymmetric sensing setting that combines high-resolution (HR) depth with ULR RGB, preserving dense geometry while restricting fine-grained visual information.
To address the severe information imbalance between HR depth and ULR RGB, we propose a joint 2D framework using HR geometry to guide semantic-oriented RGB reconstruction and RGB-D segmentation.
Despite reliable frame-level predictions, consistent scene-level understanding remains challenging under the asymmetric HR depth--ULR RGB setting.
We therefore develop an end-to-end 2D-to-3D pipeline that consolidates 2D semantic features for 3D segmentation.
Experiments on ScanNet show that our method achieves the best 2D and 3D segmentation performance among privacy-preserving approaches and delivers the strongest zero-shot transfer to SUN RGB-D and SceneNN.
Privacy recoverability analysis shows that our proposed HR depth--ULR RGB input reduces the recoverability of sensitive data, and real-robot experiments demonstrate the utility of the resulting 3D semantics for object-goal navigation.
\end{abstract}

\section{INTRODUCTION} 
As mobile robots increasingly operate in human-centered environments such as homes, offices, hospitals, and care facilities, their onboard cameras may inadvertently capture user privacy data such as faces, readable text and display contents, raising user privacy concerns~\citep{taras2023arXiv}.
Several existing approaches mitigate visual privacy risks by using 
source-level ultra-low-resolution (ULR) sensing, with resolutions such as \(16\times16\) or \(16\times12\) pixels to restrict visual details at acquisition time~\citep{Ryoo2018aaai,huang2026roman, huang2026arxiv}. 
At the same time, many robotic applications, such as mapping and navigation, benefit from explicit geometry understanding of the surrounding 3D~environment.
However, ULR RGB alone provides limited semantic detail and geometric information, constraining both accurate semantic perception and downstream tasks that rely on 3D scene understanding.

Depth sensing provides direct metric structure without capturing fine-grained visual appearance, and has therefore been explored as a privacy-preserving modality for robotic perception~\citep{li2016icisp, baselizadeh2023case, jain2024tai, liu2025iccv}. 
High-resolution (HR) depth offers dense geometry information that is naturally complementary to the coarse appearance cues retained by ULR~RGB. 
However, jointly exploiting these HR depth and ULR RGB for semantic perception remains largely unexplored. 
Recent ULR RGB semantic segmentation has established a basis for privacy-preserving robot navigation using improved semantic predictions~\citep{huang2026arxiv}.
We therefore propose to combine HR depth with source-level ULR RGB for privacy-preserving semantic segmentation. 
Specifically, RGB observations are restricted to $16\times12$ pixels, while depth is retained at $640\times480$, limiting fine-grained visual acquisition while preserving dense geometry information.

Existing RGB-D semantic segmentation methods~\citep{seichter2021icra, zhang2023tits, yin2024iclr}, however, typically assume RGB and depth observations at common spatial resolutions, making them unsuitable for our privacy-preserving asymmetric inputs.
This asymmetry is central to our setting: high-resolution depth provides dense spatial and geometric cues that are largely absent from ULR RGB, but effectively exploiting such complementary information requires explicitly accounting for the severe imbalance between the two modalities.
To this end, we present a novel joint-learning framework that leverages depth-conditioned visual reconstruction with RGB-D semantic learning, allowing HR geometry to guide the recovery of semantic information from ULR appearance.
To the best of our knowledge, we are the first to achieve semantic segmentation under privacy-preserving asymmetric inputs. 

While accurate image-level predictions provide useful semantic cues, directly fusing them across views to 3D is insufficient under our highly asymmetric HR depth--ULR RGB setting. 
We therefore develop an end-to-end 2D-to-3D pipeline that learns to consolidate the recovered frame-level semantic features into a full-scene representation for 3D semantic segmentation.
Fig.~\ref{fig:teaser} illustrates a real-world privacy-preserving setting and shows how our pipeline supports semantic object-goal navigation using only HR depth and ULR RGB.
To summarize, our contributions are the following:
\begin{itemize}
    \item We propose a novel 2D semantic segmentation framework that bridges the severe information imbalance between HR depth and ULR RGB by coupling depth-conditioned visual recovery with joint RGB-D semantic learning.
    \item We develop an end-to-end 2D-to-3D pipeline that learns to transfer and consolidate the recovered 2D~semantic features under the asymmetric HR depth--ULR RGB setting for 3D semantic segmentation.
    \item We introduce a privacy-preserving asymmetric HR depth--ULR RGB input setting, conduct a quantitative privacy evaluation of data disclosure across different inputs, and demonstrate its practical applicability through real-world robotic deployment.

\end{itemize}
Our implementation will be made open-source.

\section{Related Work}
\subsection{Privacy-Preserving Visual Perception for Mobile Robots}
A major source of privacy exposure is HR RGB sensing, which captures not only the visual cues required for robotic perception, but also user privacy information~\citep{taras2023arXiv}.

To mitigate privacy risks, several studies avoid detailed appearance sensing by relying on geometry-centric modalities.
For example, LiDAR-based approaches have been explored for privacy-sensitive human monitoring and activity recognition in robotic settings~\citep{enoki2024sensors,baselizadeh2025iccma}, while depth sensing has been used for privacy-preserving indoor mapping and human activity recognition~\citep{li2016icisp, jain2024tai}.
However, completely removing RGB also eliminates complementary appearance cues that are useful for semantic discrimination~\citep{huangarXivdepth}.

A common direction preserves conventional RGB sensing but applies privacy protection after HR RGB image acquisition, for example, through blurring, pixelation or anonymization~\citep{Hukkelas2019isvc, Orekondy2018cvpr, Wen2023iccv}. 
Although such approaches can reduce the privacy information exposed to downstream modules or external users, the original HR images typically need to be captured before protection is applied. 
Sensitive visual content may therefore still exist transiently within the sensing pipeline or processing buffers, which raise privacy risks in networked robotic systems~\citep{sasi2024jii}. 

A complementary strategy enforces privacy directly at the sensing stage by minimizing visual information acquisition. 
ULR sensing restricts spatial RGB detail at capture time, rather than first recording an HR image and sanitizing it afterward. 
Privacy-preserving activity recognition has been demonstrated using RGB inputs as low as \(16\times12\) pixels~\citep{Ryoo2018aaai},
while a recent user study identified resolutions around $16\times16$ as a favorable privacy--utility operating point~\citep{huang2026arxiv}. 
More recently, ULR imagery~($16\times16$) has also been explored for privacy-preserving 2D semantic segmentation~\citep{huang2026arxiv}. 
Although these methods reduce privacy exposure at the source, ULR RGB alone suffers from substantial loss of spatial details and appearance cues, constraining semantic accuracy and its utility for downstream robotic tasks.

These limitations motivate the asymmetric input configuration in this work: HR depth preserves fine-grained geometry structures, while ULR RGB provides coarse appearance cues. 

\subsection{Semantic Segmentation}
Under the privacy-preserving asymmetric RGB-D setting, a central question is how to recover accurate semantic segmentation from severe information imbalance.

\textbf{2D Semantic Segmentation.} At the image level, \mbox{RGB-D} semantic segmentation has been widely studied to exploit the complementary information provided by color and depth. 
Existing approaches broadly range from CNN-based feature fusion~\citep{hazirbas2016accv, park2017iccv, seichter2021icra} to adaptive cross-modal fusion with gating or attention~\citep{zhou2020accv, zhao2023neurocomputing}, and more recent transformer-based or joint RGB-D representation learning methods~\citep{zhang2023tits, yin2024iclr, yin2025cvpr}. 
However, these methods generally assume access to HR RGB observations, leaving their effectiveness unclear when RGB is severely constrained.
Recent ULR semantic segmentation work has explored semantic prediction from severely resolution-constrained RGB inputs~\citep{huang2026arxiv}, but treats RGB as the sole sensing modality. 
Existing RGB-D and ULR semantic segmentation therefore address complementary settings: the former leverages depth cues but assumes rich RGB input, whereas the latter constrains RGB resolution without exploiting geometry information. 
Their intersection, semantic segmentation from ULR RGB and HR depth, remains underexplored.

\textbf{3D Semantic Segmentation.} 
Image-space semantics are inherently view dependent and lack persistent metric localization, which limits their applicability to downstream robotic tasks.
Geometry-only 3D semantic segmentation addresses this limitation by reasoning directly over point clouds, voxels or meshes, exploiting spatial structure and neighborhood context for semantic prediction~\citep{huang2018cvpr, huang2019cvpr}. 
Despite strong geometry reasoning capabilities, geometry alone may remain ambiguous for semantically distinct categories with similar shapes or spatial configurations~\citep{huangarXivdepth}.

Appearance information can complement geometry representations. 
A common strategy directly augments 3D points or voxels with RGB attributes~\citep{park2022cvpr,lai2022cvpr}, while other methods extract learned image features and project them into 3D, demonstrating the benefit of richer 2D representations over naive color fusion~\citep{dai2018eccv,jaritz2019ICCVW}. 
Subsequent approaches further exploit cross-view evidence through learned multi-view aggregation or bidirectional 2D--3D interaction, or visual-to-3D knowledge transfer~\citep{robert2022cvpr,wu2024cvpr,carreaud2026isprsarchives}.

However, existing 2D to 3D approaches typically rely on semantic features extracted from HR RGB imagery.
Building on this asymmetric RGB-D setting, it remains unclear whether the recovered semantic representations can be effectively transferred and consolidated across views in 3D. 
We therefore aim to address this gap by lifting the 2D semantics into metric 3D for full-scene semantic segmentation.

\begin{figure}[!t]
  \makebox[\columnwidth][l]{%
    \includegraphics[width=1\columnwidth]{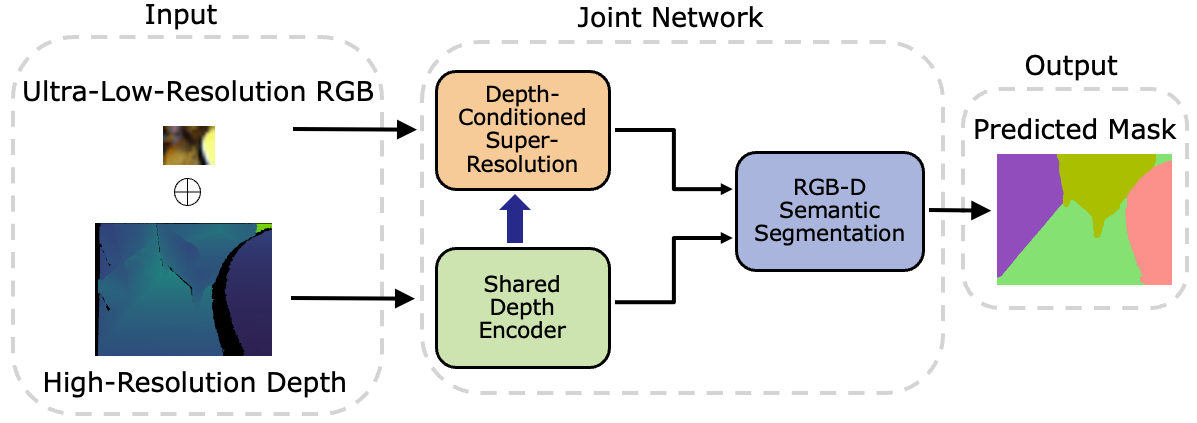}%
  }
  \caption{
  Overview of our joint 2D semantic segmentation framework.
  Given ULR RGB and HR depth inputs, a shared depth encoder extracts hierarchical geometry features that both condition the semantic-oriented super-resolution branch and support subsequent RGB-D semantic segmentation together with the reconstructed RGB representation.
}
  \label{fig:2d_framework}
  \vspace{-0.5cm} 
\end{figure}

\section{Our Approach} 
\subsection{2D Semantic Segmentation Framework}
\subsubsection{Framework Overview}
We consider a highly asymmetric RGB-D sensing configuration, where an HR depth map of \(640\times480\) is paired with a ULR RGB image of \(16\times12\). 
To address the severe information imbalance between the two modalities, we formulate super-resolution~(SR) and semantic segmentation~(SS) as a joint learning problem rather than two independent stages. 
As illustrated in Fig.~\ref{fig:2d_framework}, our framework consists of a shared depth encoder, a depth-conditioned SR branch, and an RGB-D semantic segmentation branch.
The depth encoder extracts high-resolution geometry features that condition RGB reconstruction and are also incorporated into semantic prediction.
The reconstructed appearance and depth representations are then jointly used to produce the final semantic mask.
By jointly optimizing SR and segmentation, the framework encourages the recovered RGB representation to preserve task-relevant semantic information rather than serving reconstruction alone.

\subsubsection{Shared Depth Encoder} \label{sec:depth_encoder}
Under the highly asymmetric RGB-D setting, the ULR RGB input loses substantial spatial structure, whereas the HR depth map preserves rich geometry but may contain locally unreliable measurements and irregular discontinuities. 
To effectively exploit this complementary modality, we adopt a Deformable Attention Transformer~(DAT)~\citep{xia2022cvpr} as the shared depth encoder. 
Unlike fixed-grid attention, DAT adaptively selects key–value sampling locations in a data-dependent manner, enabling the encoder to focus on informative geometry structures and aggregate context beyond locally unreliable regions.
This makes it suitable for our privacy-constrained input setting to extract reliable geometry cues from HR depth to compensate for the severe spatial information loss in ULR RGB. 
We adopt DAT-B++ as a shared depth encoder, which produces four hierarchical depth features, \(\mathcal{D}=E_D\!\left(D^{\mathrm{HR}}\right)=\left\{F_i^{d}\right\}_{i=1}^{4}\), shared by the SR and SS branches.
For SS, these features serve as stage-wise depth representations for multi-scale RGB-D feature fusion. 
For SR, they are further aggregated into a multi-scale depth-guidance pyramid \(\mathcal{G}=\{g_i^d\}_{i=1}^{K}\). 
The guidance features are spatially aligned with the corresponding SR feature maps and provide geometry information throughout the reconstruction process. 
The shared DAT encoder acts as a common geometry bridge between SR and SS, allowing HR depth cues to compensate for the limited spatial information in ULR RGB while jointly benefiting reconstruction and semantic prediction.

\subsubsection{Depth-Conditioned Semantic-Oriented SR} \label{sec:depth_sr}
To mitigate the severe information loss caused by ULR RGB inputs, we utilize a residual-in-residual dense blocks
(RRDBs)-based generator following ESRGAN~\cite{wang2018eccvworkshops} to recover a richer representation. 
The generated image serves as a task-oriented intermediate representation for semantic segmentation, while limiting perceptually recognizable privacy leakage, consistent with prior findings on reconstruction-based privacy assessment~\cite{sun2023neurips}. 
At such an extreme resolution, RGB alone provides insufficient spatial evidence for reliable reconstruction. 
We therefore exploit the geometry guidance \(\mathcal{G}\) extracted by the shared DAT encoder as introduced above. 
Inspired by Spatial Feature Transform~(SFT)~\cite{wang2018cvpr}, which modulates intermediate features through spatially varying affine parameters conditioned on external priors, we inject depth guidance into the SR features.
For an RGB feature \(F_i^{r}\) and its corresponding depth guidance \(g_i^{d}\), SFT performs spatially adaptive modulation as
\(
\left(\gamma_i^{d},\beta_i^{d}\right)
=
\phi_i\!\left(g_i^{d}\right), 
\widetilde{F}_i^{r}
=
F_i^{r}\odot\left(1+\gamma_i^{d}\right)
+
\beta_i^{d}
\), 
where \(\phi_i(\cdot)\) predicts the spatially varying scale and shift parameters from the depth guidance.
We integrate SFT into the RRDB trunk and upsampling stages, enabling geometry guidance throughout the reconstruction process.

The SR branch is supervised with a multi-scale \(\ell_1\) reconstruction objective:
$$
\mathcal{L}_{\mathrm{SR}}
=
\lambda_{\mathrm{r1}}
\left\|I^{SR}-I^{HR}\right\|_1
+
\lambda_{\mathrm{r2}}
\left\|I^{SR}_{1/2}-I^{HR}_{1/2}\right\|_1
$$
where \(\lambda_{\mathrm{r1}}\) and \(\lambda_{\mathrm{r2}}\) weight the full- and half-scale reconstruction losses respectively. The reconstructed RGB is then fed into the SS branch and jointly optimized with semantic supervision, encouraging the SR network to recover structures that are informative for downstream semantic prediction.

\subsubsection{Geometry-Guided RGB-D Semantic Segmentation}
To jointly optimize the reconstructed RGB for downstream semantic segmentation, we adopt DeepLabV3 with \mbox{ResNet-101}~\citep{chen2018eccv}. 
Although SR recovers a denser RGB representation, the extreme degradation of the ULR input leaves semantic ambiguities that cannot be reliably recovered from RGB alone. 
To further exploit HR geometry for semantic segmentation, we perform stage-wise \mbox{RGB-D} feature fusion between the multi-scale RGB features \(\{F_i^{r}\}_{i=1}^{4}\) and the corresponding depth features \(\{F_i^{d}\}_{i=1}^{4}\):
\(
F_i^{rd}
=
\mathcal{F}_i
\left(
F_i^{r},
F_i^{d},
D_i^{\mathrm{HR}}
\right),
i=1,\ldots,4
\). 
Here, \(\mathcal{F}_i(\cdot)\) denotes the geometry-guided RGB-D fusion at stage \(i\), with \(D_i^{\mathrm{HR}}\) spatially aligned to the corresponding feature resolution. The depth feature is projected and jointly rectified with the RGB feature through channel- and spatial-wise cross-modal weighting, followed by geometry-guided self-attention. 

The semantic objective combines cross-entropy with Dice, Lovász, and boundary losses:
$$
\mathcal{L}_{\mathrm{sem}}
=
\mathcal{L}_{\mathrm{CE}}
+
\lambda_{\mathrm{Dice}}\mathcal{L}_{\mathrm{Dice}}
+
\lambda_{\mathrm{Lov}}\mathcal{L}_{\mathrm{Lov}}
+
\lambda_{\mathrm{Bnd}}\mathcal{L}_{\mathrm{Bnd}}
$$
With the auxiliary segmentation head, the SS objective is
$$
\mathcal{L}_{\mathrm{SS}}
=
\mathcal{L}_{\mathrm{sem}}^{\mathrm{main}}
+
\lambda_{\mathrm{aux}}
\mathcal{L}_{\mathrm{sem}}^{\mathrm{aux}}
$$
This design allows HR geometry to complement the information-limited SR representation throughout semantic feature extraction, rather than serving merely as an additional input modality.

\subsubsection{Joint-Learning Objective}\label{sec:joint_objective}
The SR and SS branches are jointly optimized in an end-to-end manner as
\(\
\mathcal{L}_{\mathrm{joint}}
=
\mathcal{L}_{\mathrm{SR}}
+
\lambda_{\mathrm{SS}}\mathcal{L}_{\mathrm{SS}}
\).
By jointly optimizing reconstruction and semantic prediction, the SR branch is encouraged to recover structures that are beneficial to downstream segmentation. 

\begin{figure}[!t]
  \makebox[\columnwidth][l]{%
    \includegraphics[width=1\columnwidth]{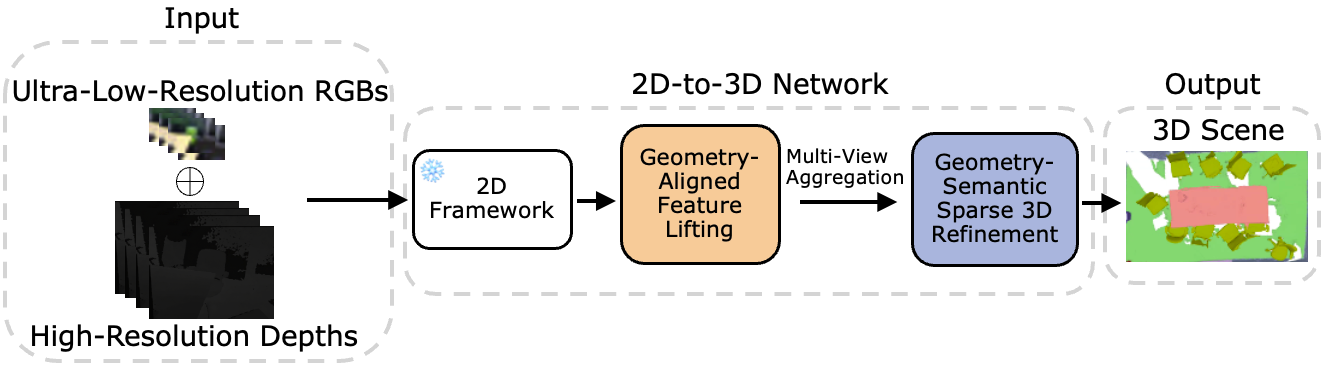}%
  }
  \caption{
Overview of our 2D-to-3D semantic segmentation pipeline.
Given multi-view ULR RGB and HR depth observations, the frozen 2D framework first extracts frame-level semantic representations.
These representations are geometrically lifted into a common 3D space, aggregated across views, and refined by a geometry-semantic sparse 3D network to produce the final 3D semantic scene prediction.
The pipeline enables spatially consistent scene-level semantic understanding under the privacy-preserving HR depth--ULR RGB sensing configuration.
}
  \label{fig:3d_pipeline}
  \vspace{-0.5cm} 
\end{figure}

\subsection{2D-To-3D Semantic Segmentation Pipeline}
\subsubsection{Pipeline Overview}
Under the privacy-preserving HR depth--ULR RGB setting, our proposed framework produces reliable frame-level semantic representations but cannot explicitly capture multi-view consistency and scene-level 3D context. 
We therefore develop an end-to-end 2D-to-3D semantic pipeline that lifts 2D semantic evidence into a unified 3D space and performs geometry-semantic reasoning with a sparse 3D network.
As illustrated in Fig.~\ref{fig:3d_pipeline}, our frozen 2D network takes the ULR RGB images and corresponding HR depth maps to extract pixel-wise semantic evidence. 
We then build an end-to-end 2D-to-3D semantic prediction pipeline that lifts this evidence into a unified 3D voxel space, aggregates observations across multiple views, and performs geometry-semantic reasoning with a learnable sparse 3D network to produce the final 3D semantic prediction. 
In this way, HR depth provides both geometry guidance for 2D perception and the spatial correspondence required to transfer semantic evidence from image space to 3D.

\subsubsection{Geometry-Aligned Feature Lifting}
For each frame \(t\), the frozen 2D model provides the pixel-wise class probability \(p_t(u)\) and the pre-classifier decoder feature \(F_t^{2D}(u)\). 
A trainable feature adapter first maps the decoder feature to a compact semantic representation
\(
f_t(u)=A_{\theta}\!\left(F_t^{2D}(u)\right).
\)
Each pixel is then back-projected into world space and associated with its canonical 3D voxel.
For each voxel, we retain the mean and variation of lifted semantic features, mean class probability, prediction uncertainty and agreement, and observation statistics. 
These 2D semantic cues are further combined with local 3D geometry, ULR color, and observation validity to form the final voxel representation.

\subsubsection{Geometry-Semantic Sparse 3D Refinement}
Although 2D lifting aligns multi-view semantic evidence in a common 3D space, the aggregated voxel representations may still contain ambiguities caused by the severely degraded ULR appearance and inconsistent observations. 
We therefore introduce a geometry-semantic 3D refinement module to explicitly reason over spatial context in 3D. 
We employ a sparse residual 3D U-Net~\citep{graham2018cvpr} to efficiently capture hierarchical geometric and semantic context while preserving fine-grained voxel structure.
Given the lifted voxel representation~\(X_i^{3D}\), the refinement network directly predicts voxel-wise semantic logits as
\(
z_i = R_{\psi}\!\left(X_i^{3D}\right)
\).
This 3D reasoning stage enables the model to refine ambiguous multi-view evidence and propagate semantic context to observed regions.

\subsubsection{3D Training Objective}
During 3D training, the pretrained 2D frontend remains frozen, while the feature adapter and sparse 3D refinement network are jointly optimized by the final 3D semantic objective. 
We use voxel-wise cross-entropy together with Lovász-Softmax loss:
$$
\mathcal{L}_{\mathrm{3D}}
=
\mathcal{L}_{\mathrm{CE}}
+
\lambda_{\mathrm{Lov}}\mathcal{L}_{\mathrm{Lov}}
$$
where \(\lambda_{\mathrm{Lov}}\) controls the contribution of the Lovász loss, complementing voxel-wise cross-entropy by improving class-level semantic segmentation quality.

\section{Experiments}

We evaluate our framework on both 2D and 3D semantic segmentation under two sensing regimes.
The \textit{non-private} setting assumes access to HR RGB together with HR depth, whereas the \textit{privacy-preserving} setting prohibits access to HR RGB and uses only ULR RGB, HR depth, or their combination.

\subsection{2D Semantic Segmentation}

\subsubsection{Datasets}
We train our 2D framework on ScanNet~\citep{dai17cvpr}, a
large-scale RGB-D benchmark of indoor scenes. Following
the official split, we use 1,201 scenes for training and 312~scenes for validation. 
For each RGB-D frame, the RGB image is cropped to \(640\times480\) and bicubically downsampled to \(16\times12\) to simulate source-level ULR RGB input following~\citep{Ryoo2018aaai}.
The corresponding depth map is retained at \(640\times480\), yielding the privacy-preserving asymmetric HR depth--ULR RGB input used throughout the 2D experiments.

\subsubsection{Baselines}
For the non-private setting, we use DFormerV2~\citep{yin2025cvpr} with HR RGB-D input as a strong \mbox{RGB-D} semantic segmentation baseline.
Under the privacy-preserving setting, we first consider two single-modality baselines:
ULRSS~\citep{huang2026arxiv} using only ULR RGB, and DeepLabV3~\citep{chen2018eccv} using only HR depth.
For the asymmetric HR depth--ULR RGB input, we further evaluate DFormerV2 under two training protocols:
(i) trained on standard HR RGB-D and directly evaluated on HR depth--ULR RGB, and
(ii) the same architecture trained using HR depth--ULR RGB.
These comparisons assess both the effect of the sensing constraint and the benefit of adapting RGB-D segmentation to the proposed privacy-preserving asymmetric input configuration.
Unless otherwise specified, all privacy-preserving baselines use the same inputs as ours, with HR depth at \(640\times480\) and/or ULR RGB at \(16\times12\).

\begin{table}[!t]
  \centering
  \setlength{\tabcolsep}{1.0pt}
  \renewcommand{\arraystretch}{1.05}
  \resizebox{\columnwidth}{!}{%
    \begin{tabular}{|c|c|c|c|c|c|c|}
      \hline
      \multirow{2}{*}{Setting}
      & \multirow{2}{*}{Input}
      & \multirow{2}{*}{Method}
      & \multicolumn{2}{c|}{ScanNet-2D}
      & \multicolumn{2}{c|}{\makecell{SUN RGB-D\\(Zero-Shot)}} \\ \cline{4-7}
      & & & mIoU & mAcc & mIoU & mAcc \\ \hline

      Non-private
      & RGB-D (HR RGB)
      & DFormerV2
      & 71.2 & 80.3
      & \textbf{50.6} & \textbf{64.5}\\ \hline

      \multirow{5}{*}{\makecell{Privacy-\\preserving}}
      & ULR RGB
      & ULRSS
      & 40.6 & 52.2
      & 17.7 & 26.3 \\ \cline{2-7}

      & HR Depth
      & DeepLabV3
      & 68.4 & 77.3
      & 37.7 & 49.9 \\ \cline{2-7}

      & \multirow{3}{*}{HR Depth + ULR RGB}
      & DFormerV2 (HR-trained)
      & 5.4 & 12.0
      & 4.1 & 9.4 \\ \cline{3-7}

      &
      & DFormerV2
      & 50.7 & 61.6
      & 25.8 & 36.4 \\ \cline{3-7}

      &
      & \textbf{Ours}
      & \textbf{72.2} & \textbf{82.1}
      & 41.2 & 56.5 \\ \hline

    \end{tabular}%
  }
  \caption{Semantic segmentation results on the ScanNet-2D validation set and SUN RGB-D test set~(zero-shot) under different sensing settings.
  As can be seen, our method achieves the strongest semantic segmentation performance on ScanNet-2D and maintains the best cross-dataset zero-shot generalization to SUN RGB-D among privacy-preserving approaches.}
  \label{tab:2d_segmentation}
  \vspace{-0.3cm}
\end{table}

\begin{figure}[!t]
  \makebox[\columnwidth]{%
    \includegraphics[width=1\columnwidth]{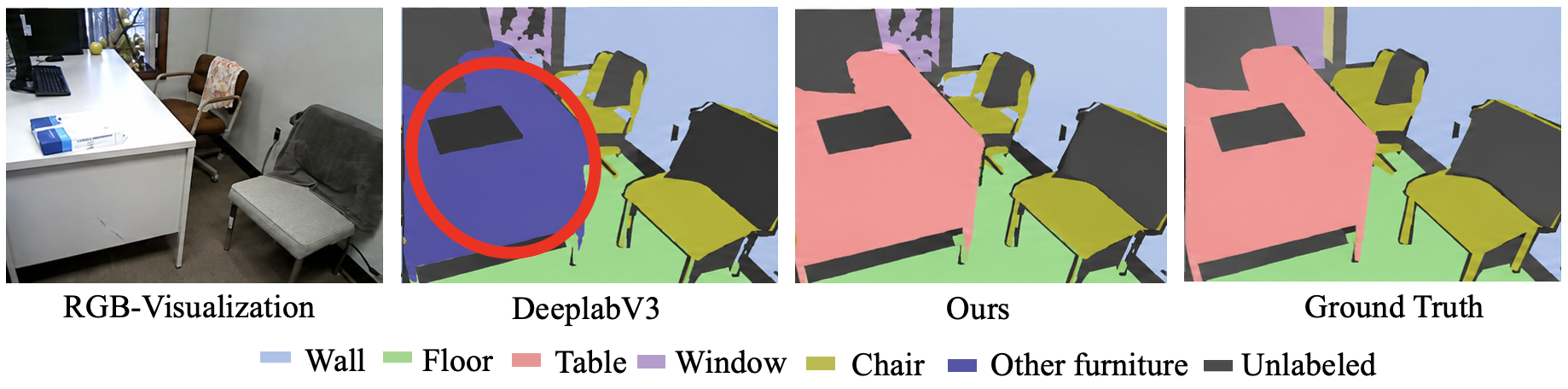}%
  }
  \caption{
 Qualitative comparison on SUN RGB-D under zero-shot transfer.
 Compared with the strongest privacy-preserving baseline, depth-only DeepLabV3, our method correctly segments the table that is otherwise confused with other furniture, showing that ULR RGB provides complementary appearance cues that further enhance the semantic utility of HR geometry.
}
  \label{fig:2d_vis}
  \vspace{-0.5cm} 
\end{figure}

\subsubsection{Implementation Details}
The SR employs $\ell_1$ reconstruction losses at the full resolution of $640\times480$ and an auxiliary half resolution of $320\times240$, weighted by  \(\lambda_{\mathrm{r1}}\): $1.0$ and  \(\lambda_{\mathrm{r2}}\): $0.5$.
For semantic supervision, we use cross-entropy with label smoothing of $0.01$, together with Dice, Lov\'asz, and boundary losses weighted by $0.05$, $0.2$, and $0.1$.
The auxiliary segmentation head is weighted by $0.4$, and the overall segmentation objective by $0.1$.
The 2D framework has an inference time of 0.37\,s per image.

\subsubsection{Evaluation Protocol and Metrics}
We evaluate in-domain performance on the ScanNet validation set.
We further evaluate the zero-shot generalization ability of our 2D framework on \mbox{SUN RGB-D} test set~\citep{song2015cvpr}, which contains 5,050 images, without fine-tuning.
To ensure consistent evaluation, SUN RGB-D labels are remapped to the 20-class ScanNet taxonomy.
We report mean Intersection over Union~(mIoU) and mean class Accuracy (mAcc).

\subsubsection{Results}
Table~\ref{tab:2d_segmentation} shows 2D semantic segmentation results on both ScanNet validation set and \mbox{SUN RGB-D} zero-shot benchmark under different input settings.

\textbf{ScanNet-2D Validation.} 
Despite operating without access to HR RGB, our method achieves performance comparable to the non-private HR RGB-D reference, demonstrating that strong semantic utility can be retained under the proposed privacy-preserving sensing configuration.
More importantly, the comparison among privacy-preserving inputs reveals the complementary roles of geometry and appearance. ULR RGB alone provides limited semantic utility, whereas HR depth offers a much stronger structural basis when fine-grained RGB appearance is unavailable.
However, simply combining the two modalities with DFormerV2 does not translate this complementarity into improved performance: even after retraining on the same asymmetric HR depth–ULR RGB input, it remains substantially below the depth-only baseline.
In contrast, our framework improves upon HR depth alone, indicating that explicitly accounting for the severe modality asymmetry enables the remaining coarse appearance cues in ULR RGB to effectively complement high-resolution geometry.

\textbf{Zero-Shot Generalization on SUN RGB-D.}
The same trend persists under cross-dataset transfer.
ULR RGB alone generalizes poorly, whereas HR depth transfers more robustly, supporting the greater cross-domain stability of geometric structure.
Our method further improves upon the depth-only baseline, showing that the coarse appearance retained by ULR RGB remains complementary even under domain shift when integrated through our asymmetric framework.
Fig.~\ref{fig:2d_vis} provides a qualitative example of this effect: the depth-only baseline confuses the table with another furniture category, while our prediction preserves the correct semantic distinction.
In contrast, DFormerV2 trained on the same HR depth–ULR RGB input remains clearly inferior, again indicating that conventional RGB-D fusion does not effectively exploit the highly imbalanced modalities.
Although a gap remains to the non-private HR RGB-D reference, our method achieves the strongest zero-shot generalization among all privacy-preserving approaches.

\subsection{3D Semantic Segmentation}
\subsubsection{Datasets}
For 3D semantic segmentation, we use the same ScanNet training and validation splits as in the 2D~experiments, with evaluation performed on the corresponding 3D scene representations.

\subsubsection{Baselines} \label{sec: 3d_baselines}
For the non-private setting, we use PTv3~\citep{wu24cvpr} on the 3D scene representation constructed from HR RGB-D observations as a strong 3D semantic segmentation baseline.
For privacy-preserving evaluation, the 2D~baselines are extended to 3D through a common multi-view fusion procedure: per-view semantic probabilities are projected onto visible mesh vertices and aggregated across views according to prediction confidence, depth consistency, and viewing angle.
This procedure is applied to ULRSS, depth-only DeepLabV3, and both DFormerV2 variants.
We additionally train PTv3 directly on the 3D representation constructed under the HR depth--ULR RGB sensing setting, providing an end-to-end 3D baseline under the same privacy constraint.

\begin{table}[!t]
  \centering
  \setlength{\tabcolsep}{1.0pt}
  \renewcommand{\arraystretch}{1.05}
  \resizebox{\columnwidth}{!}{%
    \begin{tabular}{|c|c|c|c|c|c|c|}
      \hline
      \multirow{2}{*}{Setting}
      & \multirow{2}{*}{Input}
      & \multirow{2}{*}{Method}
      & \multicolumn{2}{c|}{ScanNet-3D}
      & \multicolumn{2}{c|}{\makecell{SceneNN\\(Zero-Shot)}} \\ \cline{4-7}
      & & & mIoU & mAcc & mIoU & mAcc \\ \hline

      Non-private
      & RGB-D (HR RGB)
      & \textbf{PTv3}
      & \textbf{77.5}
      & \textbf{84.9}
      & 51.2
      & 68.0 \\ \hline

      \multirow{6}{*}{\makecell{Privacy-\\preserving}}
      & ULR RGB
      & ULRSS - 3D
      & 40.6
      & 49.8
      & 21.8
      & 30.6 \\ \cline{2-7}

      & HR Depth
      & DeepLabV3 - 3D
      & 66.3
      & 74.2
      & 47.2
      & 61.3 \\ \cline{2-7}

      & \multirow{4}{*}{HR Depth + ULR RGB}
      & DFormerV2 (HR-trained) - 3D
      & 4.8
      & 11.7
      & 5.8
      & 13.8 \\ \cline{3-7}

      &
      & DFormerV2 - 3D
      & 50.8
      & 60.1
      & 32.9
      & 44.1 \\ \cline{3-7}

      &
      & PTv3
      & 70.0
      & 78.5
      & 49.6
      & 67.1 \\ \cline{3-7}

      &
      & \textbf{Ours}
      & 72.3
      & 81.6
      & \textbf{52.3}
      & \textbf{69.5} \\ \hline

    \end{tabular}%
  }
  \caption{3D semantic segmentation performance on \mbox{ScanNet-3D} and zero-shot evaluation on SceneNN under different sensing settings.
  As can be seen, our method achieves the strongest segmentation performance among the privacy-preserving approaches on \mbox{ScanNet-3D} and maintains the best zero-shot generalization to SceneNN.}
  \label{tab:3d_segmentation}
  \vspace{-0.3cm}
\end{table}

\begin{figure}[!t]
  \makebox[\columnwidth]{%
    \includegraphics[width=1\columnwidth]{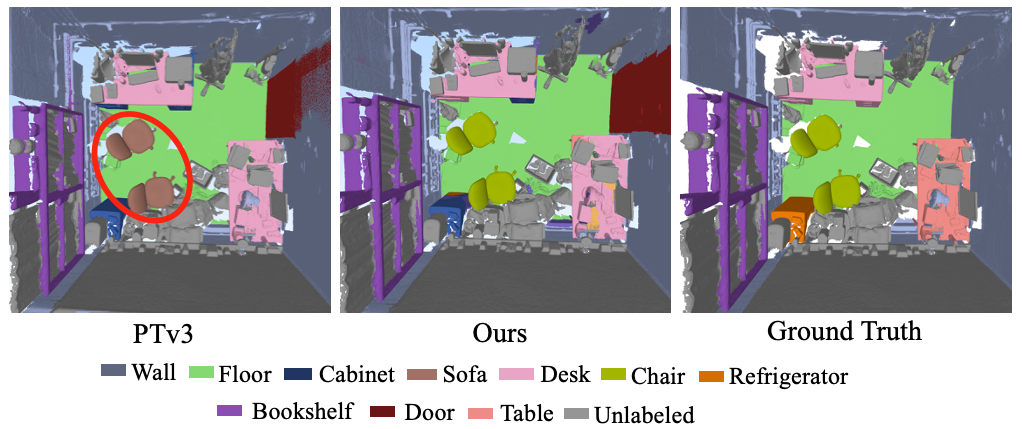}%
  }
  \caption{
Qualitative comparison on SceneNN under zero-shot transfer.
Compared with the strongest privacy-preserving baseline PTv3, which misclassifies a chair as sofa, our method preserves the correct semantic category, illustrating the benefit of incorporating lifted 2D semantic evidence into scene-level 3D reasoning.
}
  \label{fig:3d_vis}
  \vspace{-0.5cm} 
\end{figure}

\subsubsection{Implementation Details}
We employ mixed-precision training, and clip the global gradient norm to 1.0.
The 3D pipeline has a scene-level inference time of approximately 1.4\,s per scene.

\subsubsection{Evaluation Protocol and Metrics}
We evaluate in-domain 3D semantic segmentation on the ScanNet validation set.
To assess cross-dataset generalization, we further perform zero-shot evaluation on SceneNN~\citep{hua2016threedv} without fine-tuning.
We follow prior cross-dataset evaluation protocols of~\citep{huang2021supervoxel} on SceneNN. 
After filtering scenes with incomplete RGB-D data, we retain 46 scenes for evaluation.
Since 17 SceneNN categories overlap with the ScanNet 20-class taxonomy, evaluation is performed over these shared classes.
We report 3D mIoU and 3D mAcc over mesh vertices.

\subsubsection{Results}
Table~\ref{tab:3d_segmentation} reports 3D semantic segmentation performance on ScanNet-3D and zero-shot generalization to SceneNN under different sensing settings.

\textbf{ScanNet-3D Validation.} 
The non-private PTv3 baseline with access to HR RGB provides the strongest reference performance. 
Despite operating without HR RGB, our method retains strong 3D semantic segmentation performance and is close to this non-private reference. 
Among the privacy-preserving baselines, HR depth provides substantially stronger 3D semantics than ULR RGB alone, highlighting the importance of geometry information for scene-level segmentation.
For the combined HR depth--ULR RGB input, retraining DFormerV2 mitigates the input mismatch of directly using HR-trained DFormerV2, but still yields limited 3D performance, indicating that post-hoc \mbox{2D-to-3D} fusion is insufficient under this highly asymmetric setting.
Direct 3D learning with PTv3 provides a considerably stronger baseline, while our end-to-end 2D-to-3D pipeline achieves the best performance under the privacy-preserving asymmetric input setting by learning how to transfer and consolidate the recovered 2D semantic features in 3D.

\begin{table}[!t]
  \centering
  \setlength{\tabcolsep}{2.5pt}
  \renewcommand{\arraystretch}{1.05}
  \resizebox{0.5\columnwidth}{!}{%
    \begin{tabular}{|c|c|c|}
      \hline
      \multirow{2}{*}{Method}
      & \multicolumn{2}{c|}{ScanNet-3D} \\ \cline{2-3}
      & mIoU & mAcc \\ \hline

      Ours-w/o Feature Lifting
      & 70.3
      & 80.5 \\ \hline

      Ours-w/o 3D Refinement
      & 59.8
      & 68.8 \\ \hline

      Ours-w/o Both
      & 66.6
      & 77.1 \\ \hline

      \textbf{Ours}
      & \textbf{72.3}
      & \textbf{81.6} \\ \hline
    \end{tabular}%
  }
  \caption{Ablation study on the ScanNet-3D validation set.
  Our proposed full model achieves the best 3D semantic segmentation performance, confirming the effectiveness of both components.}
  \label{tab:ablation_scannet3d}
  \vspace{-0.65cm}
\end{table}

\textbf{Zero-Shot Generalization on SceneNN.} 
We further evaluate cross-dataset generalization by directly transferring all trained models to SceneNN without fine-tuning.
Under the privacy-preserving setting, HR depth generalizes better than ULR RGB alone, indicating that geometry structure provides a more stable cue across indoor domains under the privacy-preserving setting.
For the combined HR depth--ULR RGB input, 2D-to-3D fusion baselines remain sensitive to domain shift.
In contrast, direct learned 3D reasoning with PTv3 yields a large improvement compared to non-end-to-end baselines.
Fig.~\ref{fig:3d_vis} further illustrates local semantic ambiguities in PTv3, such as confusing a chair with a sofa, whereas our method preserves the correct category.
By explicitly lifting 2D semantic representations and integrating them with geometric context in 3D, our framework achieves the best zero-shot performance among the privacy-preserving approaches.

\textbf{Ablation Study.} 
Table~\ref{tab:ablation_scannet3d} evaluates the contribution of each component in our pipeline.
\textit{Ours-w/o Feature Lifting} removes the learned 2D-to-3D feature lifting, while \textit{Ours-w/o 3D Refinement} removes the 3D refinement module.
\textit{Ours-w/o Both} uses a non-end-to-end pipeline, where frame-level 2D predictions are fused into 3D following the procedure in Sec.~\ref{sec: 3d_baselines}.
Removing either feature lifting or 3D refinement degrades performance.
Notably, feature lifting without 3D refinement performs even worse than the multi-view fusion baseline, indicating that the transferred 2D features require learned 3D context to resolve cross-view inconsistency and local ambiguity. 
Our full model achieves the best performance, demonstrating the complementary roles of 2D feature transfer and 3D refinement.

\subsection{Privacy Recoverability Analysis} \label{sec:privacy_analysis}
Following~\citep{huangarXivdepth}, we assess potential data disclosure by analyzing the recoverability of sensitive information from each sensing modality.
We collect a small dataset of 80 images covering four privacy-sensitive categories: credit card, passport, private chat, and face, with 20 images per category.
Text recoverability is evaluated on the first three categories (60 images), while appearance recoverability is evaluated on the passport and face subsets (40 images).
We compare five different inputs: HR RGB, HR depth, ULR RGB, matched SR and shuffled SR.
Matched SR is reconstructed using the corresponding input pair, whereas shuffled SR uses a mismatched pair from another sample.

\subsubsection{Sensitive Text Recoverability}
We use \texttt{PP-OCRv6\_medium}~\citep{zhang2026arxiv} with a confidence threshold of 0.95 and evaluate both exact text recovery and character-level similarity.
As shown in Table~\ref{tab:privacy_text}, HR RGB retains substantial readable text, with 331 of 384 detected text instances recovered correctly.
In contrast, HR depth, ULR RGB, and both SR variants recover no text exactly.
Their character-level similarities also remain negligible, and the small gap between matched and shuffled SR indicates that reconstruction does not meaningfully recover sample-specific textual information.

\subsubsection{Appearance Attribute Recoverability}
We further follow~\citep{huangarXivdepth} and use Qwen3-VL-8B-Instruct~\citep{bai2025arXiv} to predict privacy-sensitive appearance attributes, including hair color, iris color, and eyeglass presence.
We report the macro accuracy across the three attributes.
As shown in Table~\ref{tab:privacy_appearance}, HR RGB retains strong appearance information, achieving 95.0\% accuracy, whereas both HR depth and ULR RGB yield 0.0\%.
The marginal difference between matched and shuffled SR indicates that the reconstruction recovers little sample-specific appearance information related to the true privacy attributes.
These results indicate that the proposed HR depth--ULR RGB input substantially reduces recoverability of both textual and appearance information relative to HR RGB sensing.

\begin{table}[!t]
  \centering
  \setlength{\tabcolsep}{2.5pt}
  \renewcommand{\arraystretch}{1.05}
  \captionsetup[subtable]{labelfont=normalfont,textfont=normalfont}

  \begin{subtable}{\columnwidth}
    \centering
    \caption{Sensitive Text Recoverability}
    \label{tab:privacy_text}
    \resizebox{\columnwidth}{!}{%
      \begin{tabular}{|c|c|c|c|c|c|}
        \hline
        Metric
        & HR RGB
        & HR Depth
        & ULR RGB
        & Matched SR
        & Shuffled SR \\ \hline

        Exact recovery
        & 331/384
        & 0/384
        & 0/384
        & 0/384
        & 0/384 \\ \hline

        Character similarity
        & 100.0\%
        & 0.0\%
        & 0.91\%
        & 0.48\%
        & 0.39\% \\ \hline
      \end{tabular}%
    }
  \end{subtable}

  \vspace{1mm}

  \begin{subtable}{\columnwidth}
    \centering
    \caption{Appearance Attribute Recoverability}
    \label{tab:privacy_appearance}
    \resizebox{\columnwidth}{!}{%
      \begin{tabular}{|c|c|c|c|c|c|}
        \hline
        Metric
        & HR RGB
        & HR Depth
        & ULR RGB
        & Matched SR
        & Shuffled SR \\ \hline

        Macro accuracy
        & 95.0\%
        & 0.0\%
        & 0.0\%
        & 10.5\%
        & 9.9\% \\ \hline
      \end{tabular}%
    }
  \end{subtable}

  \caption{Privacy-sensitive information recoverability analysis.}
  \label{tab:privacy_recoverability}
\end{table}

\begin{table}[!t]
  \centering
  \setlength{\tabcolsep}{1.0pt}
  \renewcommand{\arraystretch}{1.05}
  \resizebox{\columnwidth}{!}{%
    \begin{tabular}{|c|c|c|c|c|c|c|c|}
      \hline
      Method
      & Chair
      & Sofa
      & Table
      & Door
      & Total
      & Grounding Precision
      & Grounding Recall \\ \hline

      GT Objects
      & 2
      & 1
      & 2
      & 1
      & 6
      & --
      & -- \\ \hline

      Ours
      & 1/1
      & 1/1
      & 2/4
      & 1/1
      & 5/7
      & 71.4\%
      & 83.3\% \\ \hline

    \end{tabular}%
  }
  \caption{Real-robot semantic object-goal navigation results.
  GT Objects denotes the number of target object instances in the scene, while entries for our method report valid goal candidates over generated candidates.
  Grounding Precision measures the fraction of generated candidates that correspond to valid goals, and Grounding Recall measures the fraction of ground-truth objects successfully covered by at least one valid goal candidate.}
  \label{tab:real_robot_goal_grounding}
  \vspace{-0.65cm}
\end{table}

\subsection{Semantic Object-Goal Navigation Case Study}
\subsubsection{Task and Evaluation Protocol}
We further evaluate the practical utility of our 3D pipeline on semantic object-goal navigation, where a robot is required to approach a specified object~\citep{sun2024tase}.
Experiments are conducted in a real-world privacy-constrained indoor environment shown in Fig.~\ref{fig:teaser}, using the 3D semantic map constructed from HR depth and ULR RGB without access to HR RGB observations.
We consider four target objects: \textit{chair}, \textit{sofa}, \textit{table}, and \textit{door}.

For each semantic query, the predicted 3D regions of the target class are projected onto the navigation map to generate candidate goal poses.
A navigation attempt is considered valid if its resulting stopping pose lies within the valid goal region of a corresponding ground-truth object.
We report \emph{Grounding Precision}, defined as the number of valid goal candidates divided by the total number of generated candidates, and \emph{Grounding Recall}, defined as the number of ground-truth objects covered by at least one valid candidate divided by the total number of ground-truth objects.

\subsubsection{Results}
As shown in Table~\ref{tab:real_robot_goal_grounding}, our pipeline achieves a Grounding Precision of \(71.4\%\) and a Grounding Recall of \(83.3\%\).
Among the six ground-truth objects, five are successfully grounded; the only missed instance is one of the two chairs.
Both table instances are successfully identified, while additional noisy fragments produce two fake goal candidates and account for the lower precision.
The sofa and door are both grounded without additional candidates.
These results demonstrate that the 3D semantic representation obtained from the asymmetric HR depth--ULR RGB input can provide reliable and actionable semantic goals for real-world robot navigation under privacy constraints.
For additional details and experiments, please refer to the accompanying video.

\section{Conclusion}
This work introduced a privacy-preserving asymmetric sensing framework that combines HR depth with ULR RGB for 2D-to-3D semantic segmentation.
The results suggest several design considerations for privacy-constrained perception.
The strong asymmetry between HR depth and ULR RGB should be explicitly accounted for, with geometry providing the primary structural information and ULR appearance serving as a complementary cue.
Our privacy analysis further suggests that task-relevant semantic recovery can be partially decoupled from the recovery of sensitive visual details, supporting reconstruction as a task-oriented intermediate representation.
Future work will validate this setting with dedicated ULR imaging hardware under real sensor characteristics.

\bibliographystyle{IEEEtranSN}
\bstctlcite{IEEEexample:BSTcontrol}
\footnotesize
\balance
\bibliography{icra2027}

\end{document}